\documentclass{article}
\usepackage{ijcai20}

\usepackage{times}
\usepackage{soul}
\usepackage{url}
\usepackage[hidelinks]{hyperref}
\usepackage[utf8]{inputenc}
\usepackage[small]{caption}
\usepackage{graphicx}
\usepackage{amsmath}
\usepackage{amsfonts}       
\usepackage{amsthm}
\usepackage{booktabs}
\usepackage{algorithm}
\usepackage{algorithmic}

\usepackage{color}
\usepackage{subfigure}
\title{CDC: Classification Driven Compression for Bandwidth Efficient Edge-Cloud Collaborative Deep Learning}

\author{
Yuanrui Dong$^1$\and
Peng Zhao$^1$\and
Hanqiao Yu$^1$\and
Cong Zhao$^2$\footnote{Corresponding Author}\And
Shusen Yang$^1$\\
\affiliations
$^1$National Engineering Laboratory for Big Data Analytics, Xi’an Jiaotong University, China\\
$^2$Department of Computing, Imperial College London, UK\\
\emails
d2484769248@stu.xjtu.edu.cn,
p.zhao@mail.xjtu.edu.cn,
yuhanqiao@stu.xjtu.edu.cn,
c.zhao@imperial.ac.uk,
shusenyang@mail.xjtu.edu.cn
}

\begin{document}

\maketitle

\begin{abstract}
  The emerging edge-cloud collaborative Deep Learning (DL) paradigm aims at improving the performance of practical DL implementations in terms of cloud bandwidth consumption, response latency, and data privacy preservation.
  Focusing on bandwidth efficient edge-cloud collaborative training of DNN-based classifiers, we present CDC, a Classification Driven Compression framework that reduces bandwidth consumption while preserving classification accuracy of edge-cloud collaborative DL.
  Specifically, to reduce bandwidth consumption, for resource-limited edge servers, we develop a lightweight autoencoder with a classification guidance for compression with classification driven feature preservation, which allows edges to only upload the latent code of raw data for accurate global training on the Cloud.
  Additionally, we design an adjustable quantization scheme adaptively pursuing the tradeoff between bandwidth consumption and classification accuracy under different network conditions, where only fine-tuning is required for rapid compression ratio adjustment.
  Results of extensive experiments demonstrate that, compared with DNN training with raw data, CDC consumes 14.9$\times$ less bandwidth with an accuracy loss no more than 1.06\%, and compared with DNN training with data compressed by AE without guidance, CDC introduces at least 100\% lower accuracy loss.
\end{abstract}

\section{Introduction} \label{sec:introduction}
Recently, the emerging paradigm of collaborative Edge Intelligence (EI) rapidly draws significant interests from both academia \cite{8736011,8763885} and industry\cite{googleEI,msEI}, where the deployment of various Artificial Intelligence (AI) applications is pushed from mega-scale cloud datacenters to heterogeneous devices at network edges closer to explosive end data \cite{networking2016cisco}.
Confronting \textit{intensive computations on the Cloud} and \textit{expensive end data uploading} that are inherently controversial in conventional cloud-based AI deployments, EI has been demonstrated to be promising in reducing cloud bandwidth consumption and response latency, as well as preserving data privacy \cite{8736011,8763885}.

As the frontier of the latest flourish of AI, Deep Learning (DL) is of extraordinary success.
Even so, the EI-driven solution above is indisputably appealing but challenging to the deployment of urgently required DL applications.
Conventionally, highly accurate DL models can be centrally trained on the Cloud with abundant resources~\cite{mell2011nist}, which, however, requires enormous raw data (\emph{e.g.} images and videos) to be uploaded, and induces expensive bandwidth consumptions~\cite{shi2016promise}, especially considering skyrocketing end data sources (\emph{e.g.} IoT devices).
Alternatively, direct DNN training at edges with easily accessible end data can significantly reduce the cloud bandwidth consumption~\cite{satyanarayanan2017emergence}, but, in practice, it is quite difficult (or even infeasible) to train naturally complex highly accurate models under severe edge resource constraints.
Therefore, the \textit{tradeoff between cloud bandwidth consumption and model accuracy} has to be explicitly addressed for effective edge-cloud collaborative DL.

Obviously, one of the most intuitive methods to reduce cloud bandwidth consumption is to compress raw data at edges before uploading them to the Cloud.
Generally, Lossy compression~\cite{Shaham2018Deformation} manages to obviously reduce the data size at the cost of losing details, where the accuracy of models trained on such compressed data may be severely impacted.
Therefore, it is critical to develop a data compression method that can obviously reduce the data size while selectively preserving valuable details for accurate model training.

Focusing on such an issue, treating DNN-based classification as a use case, we present a Classification Driven Compression (CDC) framework to reduce bandwidth consumption while preserving model accuracy for effective edge-cloud collaborative DNN training.
Particularly, for resource limited edge servers, we design a lightweight autoencoder (AE) with a classification guidance for compression with classification driven feature preservation.
In addition, we develop an adjustable quantization scheme for the tradeoff between bandwidth consumption and classification accuracy, where only fine-tuning is required for rapid compression ratio adjustment.
Contributions of this paper are as follows:

\begin{enumerate}
  \item
  We propose CDC for bandwidth efficient edge-cloud collaborative DNN classifier training, where raw data remain at edges, and only small size latent codes are uploaded to the Cloud.
  To the best of our knowledge, CDC is the first classification driven compression method.
  \item
  We present a classification guidance approach for the edge compression AE to selectively learn representative features for accurate classification, which allows CDC to reduce bandwidth consumption and preserve classification accuracy simultaneously.
  Besides, an adjustable quantization scheme is developed to rapidly achieve the tradeoff between bandwidth consumption and classification accuracy under different network conditions.
  \item
  We conducted extensive experiments to evaluate CDC's performance, compared with collaborative DNN training with raw data, CDC manages to consume 14.9$\times$ less bandwidth while introducing an accuracy loss no more than 1.06\%.
  Besides, compared with DNN training based on data compressed by traditional autoencoder, CDC manages to introduce at least 100\% lower accuracy loss with the same compression ratio.
\end{enumerate}

\section{Related Work} \label{sec:relatedwork}

\paragraph{Deep learning with edge computing.}
Emerging efforts \cite{8736011,8763885} have been demonstrating that deploying DL models at network edges can bring about significant performance gains in terms of cloud bandwidth saving, inference latency reduction, and privacy preservation, comparing to predominating cloud-based approaches.
For instance, \cite{teerapittayanon2017distributed,li2018learning} use early-exit at a proper intermediate DNN layer to reduce inference latency, and \cite{li2018edge} focuses on online optimization of the exit point.
However, in these efforts, DL models are trained on the Cloud with expensive bandwidth consumptions.
Some researchers are interested in distributed training and focus on issues caused by model aggregation, \emph{e.g.} bandwidth consumption~\cite{hardy2017distributed,tang2018communication,smith2017federated}, training latency~\cite{tang2018communication}, stragglers, and fault tolerance~\cite{smith2017federated}.
However, these efforts do not leverage edge and cloud resources collaboratively.

\paragraph{Autoencoder with auxiliary task.}
The potential of autoencoder in image compression has already been widely demonstrated~\cite{theis2017lossy,zhou2018variational,li2018learning_}, and many of these are comparable to the best image compression standards in terms of perceptual metrics.
However, these efforts are not for resource-constrained edges.
Moreover, in the field of multi-task learning, auxiliary tasks have also been explored through the use of hints for neural networks, and the autoencoder is used as an auxiliary task to improve the performance of classification tasks~\cite{liu2016algorithm,le2018supervised}.
Beyond that, autoencoder can also accept multiple loss functions, as a main task, to learn a more efficient model~\cite{cipolla2018multi}.
However, these efforts do not focus on compressing data while retaining features that are valuable for the auxiliary task.

\section{Classification Driven Compression for Edge-cloud Collaborative DNN Training} \label{sec:solution}
\subsection{Basic Idea} \label{subsec:basicidea}
\begin{figure}[tbp]
	\centering
	\includegraphics[width=1\linewidth]{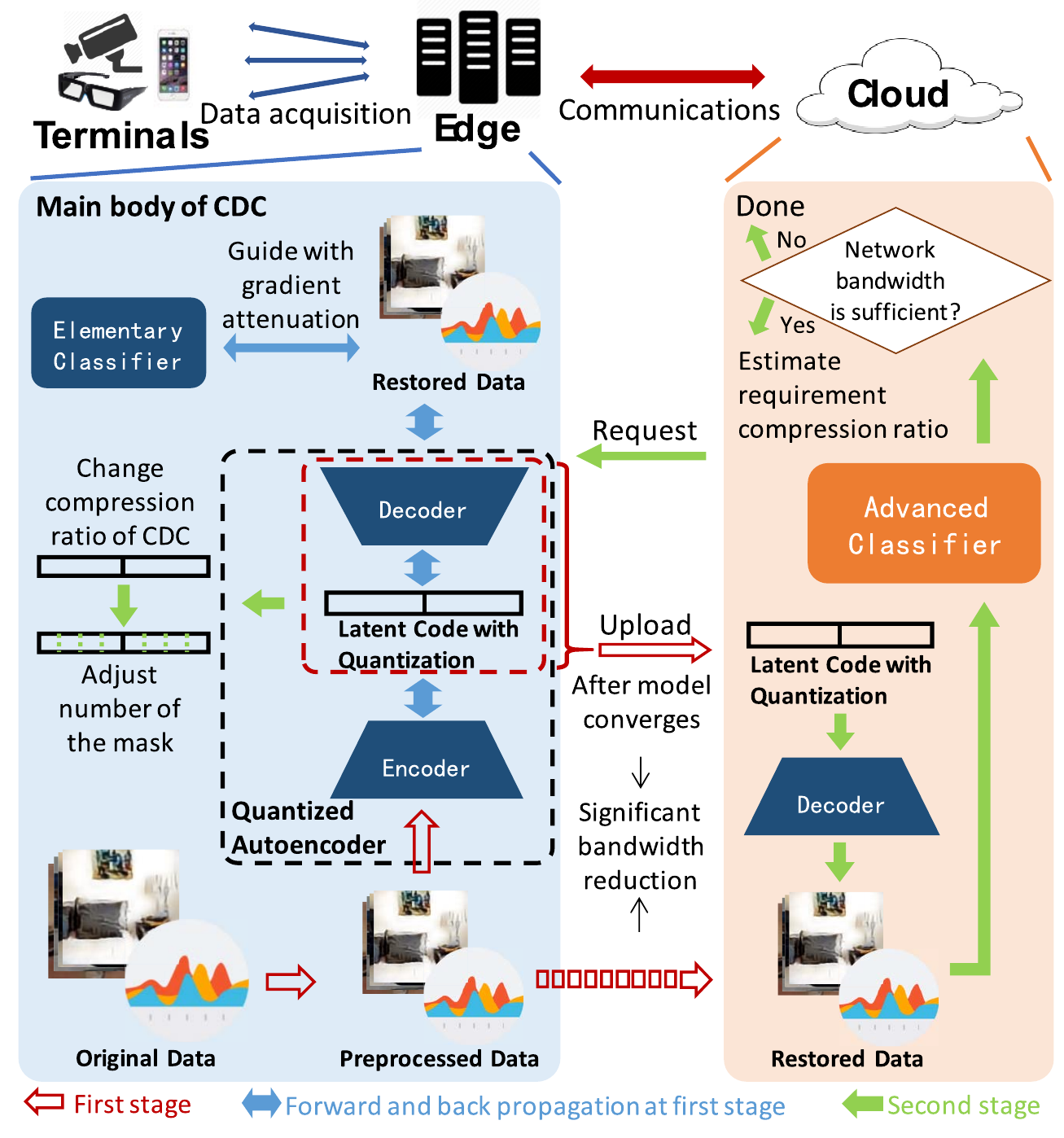}
  \caption{Edge-cloud collaborative DNN training framework. Left: On the Edge. Right: On the Cloud.}
  \label{framework-pic}
\end{figure}

In this paper, we treat the CNN-based classification as a use case, and focus on the problem of raw data compression that reduces the bandwidth consumption and preserves the classification accuracy simultaneously for effective edge-cloud collaborative DNN training.
The architecture of our collaborative DNN training framework is shown in Figure~\ref{framework-pic}.

In our system, the edge server acquires labelled data from various kinds of terminal devices, and a high-accuracy \textit{advanced classifier} (AC, \emph{e.g.} ResNet, ResNeXt, DPN) on the Cloud requires to be trained.
Different with existing approaches of large scale DNN training on the Cloud, raw data reside on the edge, where a CDC process is conducted to reduce the bandwidth consumption for the training of AC.
In particular, CDC on the edge is based on a \textit{quantized autoencoder} (AE) and an \textit{elementary classifier} (EC).
Here, the quantized AE is responsible for compressing raw data into representative features (\emph{i.e.} latent codes) with significantly reduced sizes, where an \textit{adjustable quantizer} is adopted to achieve the tradeoff between data compression ratio and feature preservation ratio.
Then, EC focuses on the similar classification task with that of AC, which has a lower accuracy but can be deployed on resource limited edge servers.
Integrating EC into CDC aims at providing a `guidance' for the construction of AE on the edge, which allows AE to effectively compress raw data while preserving information that is critical to the performance of AC on the Cloud.
The workflow of our system can be divided into two stages:

\textbf{Stage 1:}
For the edge server, AE and EC are jointly trained through gradient descent using local raw data.
As shown in Figure~\ref{framework-pic}, the forward and backward propagation of the training loss is illustrated with blue arrows.
Specifically, the training of EC is based on the classification loss between restored data from AE and corresponding raw labels.
AE is trained with the joint loss of both the reconstruction loss of itself and an attenuated classification loss of EC (which is treated as the `classification guidance').
After the convergence of AE and EC, raw data are compressed into latent codes, which are uploaded with corresponding labels together with the decoder to the Cloud for the training of AC.

\textbf{Stage 2:}
Receiving latent codes and the decoder from the edge server, AC is trained based on restored data on the Cloud.
When AC converges, the edge-cloud bandwidth is evaluated: if it is intensely occupied, the training of AC is terminated;
else if the bandwidth is sufficient, requests of clearer data are sent to the edge server, together with a proper edge compression ratio estimated by the Cloud.
Receiving the request from the Cloud, the edge adjusts its compression ratio through the quantizer of AE and re-enters Stage 1.

\subsection{Classification-guided Autoencoder}
\label{Guidance}
\subsubsection{Overview}
As mentioned above, we construct an autoencoder on the edge to compress raw data for the reduction of bandwidth consumption of AC training on the Cloud.
To preserve the accuracy of AC trained based on compressed data (\emph{i.e.} latent codes, labels and the decoder), AE and EC are jointly trained on the edge, where the classification loss of EC is used to guide the training of AE.
Specifically, the training of EC is a \textit{supervised auxiliary task} that helps the training of AE focusing more on retaining critical features for accurate classifications.
In fact, such a classification guidance can be viewed as a certain form of inductive transfer that helps to introduce certain preferences into model training (\emph{i.e.} the successive classification on the Cloud in this paper) by introducing an inductive bias (\emph{i.e.} the classification loss of EC in this paper).

\subsubsection{Classification Loss of EC and Reconstruction Loss of AE}
Under a supervised learning setting, EC's basic target is to learn a mapping from a vector of inputs
$\mathbf{x}\in \mathbb{R}^{N}$
to a vector of labels
$\mathbf{y}\in \mathbb{R}^{C}$
with the minimum \textit{classification loss} $L_{C}$, which is iteratively optimized through the training based on a batch of raw data
$\left ( \mathbf{x}_{1},\mathbf{y}_{1} \right ),\ldots ,\left ( \mathbf{x}_{t},\mathbf{y}_{t} \right )$.
The classifier output layer has a supervised loss:
\begin{equation} \label{eq:ecloss}
  L_{C}\left ( h_{C}\left (  \mathbf{x} \right ),\mathbf{y} \right ) = -\frac{1}{t}\sum_{i=1}^{t}\sum_{j=1}^{C}\mathbf{y}_{i}^{(j)} \ln h_{C}^{(j)}(\mathbf{x}_{i}),
\end{equation}
where $h_{C}\left ( \mathbf{x} \right )$ denotes the output of the classifier.

Under the same setting, AE's basic target is to learn a mapping from a vector of inputs
$\mathbf{x}\in \mathbb{R}^{N}$
to a vector of reconstructed outputs
$\mathbf{x}'\in \mathbb{R}^{N}$
with the minimum \textit{reconstruction loss} $L_{A}$, which is also iteratively optimized through the training based on i.i.d. raw data batches.
For a compression AE, the output layer has a reconstruction loss:
\begin{equation} \label{eq:aeloss}
  L_{A}\left ( h_{A}\left ( \mathbf{x} \right ),\mathbf{x} \right ) = \frac{1}{2t}\sum_{i=1}^{t}\left \|h_{A}\left ( \mathbf{x}_{i} \right )-\mathbf{x}_{i}  \right \|_{2}^{2},
\end{equation}
where $h_{A}\left ( \mathbf{x} \right )$ denotes the output of AE.

In our system, the output of AE is directly fed into EC, and we define the \textit{full loss} as:
\begin{equation} \label{eq:1}
  L_{F} = L_{A}\left ( h_{A}\left ( \mathbf{x} \right ),\mathbf{x} \right )  + L_{C}\left ( h_{C}\left ( h_{A}\left ( \mathbf{x} \right ) \right ),\mathbf{y} \right ).
\end{equation}

As for the rationality of DNN training based on the full loss above, Le \emph{et al.}~\cite{le2018supervised} has proved that a classifier trained with a loss function similar to Equation~(\ref{eq:1}) can perform better and be effectively regularized under certain assumptions.
It indicates that feature extractions of both the classifier and the AE are compatible to a certain extent, and it is reasonable to train the two models jointly for data compression that preserves classification-related features.

\subsubsection{Prioritized Joint Training of AE and EC}
For the application of massive amounts of practical data, relatively deeper nonlinear networks with more complex structures (that can still be deployed on resource-limited edges) should be adopted.
In this case, we have an observation that AE directly trained with the full loss $L_{F}$ in Equation~(\ref{eq:1}) may not converge.
To address this issue, we design a parameter, the \textit{classification attenuation rate} $\alpha \geq 1$, to adjust the priority of $L_{C}$ and $L_{A}$.
Specifically, the full loss is modified as:
\begin{equation} \label{eq:3}
 L_{F} = L_{A}\left ( h_{A}\left ( \mathbf{x} \right ),\mathbf{x} \right )  + \frac{1}{\alpha} L_{C}\left ( h_{C}\left ( h_{A}\left ( \mathbf{x} \right ) \right ),\mathbf{y} \right ).
\end{equation}
With such a full loss, the impact of classification guidance can be controlled by adjusting $\alpha$ to achieve the convergence of AE.
When $\alpha$ is large enough, the impact of classification guidance is negligible.

In our system, AE and EC are jointly trained through gradient decent with the same learning rate but different losses:
the gradient of EC training is calculated as:
\begin{equation} \label{eq:ecgradient}
    \delta^{EC} = \nabla L_{C}\left ( h_{C}\left ( h_{A}\left ( \mathbf{x} \right ) \right ),\mathbf{y} \right ),
\end{equation}
and the gradient of AE training is calculated as:
\begin{equation} \label{eq:4}
    \delta^{AE} = \nabla L_{A}\left ( h_{A}\left ( \mathbf{x} \right ),\mathbf{x} \right ) + \frac{1}{\alpha} \nabla L_{C}\left ( h_{C}\left ( h_{A}\left ( \mathbf{x} \right ) \right ),\mathbf{y} \right ).
\end{equation}

Based on the above design, the impact of classification guidance on the training of AE can be controlled by adjusting the attenuation rate $\alpha$:
a lower $\alpha$ will preserve more critical data features for high-accuracy classification in the constructed AE, which, however, will lead to a more obvious impact on the convergence of AE.
Besides, attenuation rate $\alpha$ also significantly affects the visual result of restored data.

\subsection{Compression Based on Quantization}
\label{Quantization}
To achieve the tradeoff between bandwidth consumption and classification accuracy in edge-cloud collaborative DNN training, we construct a compression autoencoder with an adjustable quantizer for the edge server.
In particular, based on the stochastic quantization method proposed by Theis \emph{et al.}~\cite{theis2017lossy}, we design a lightweight compression autoencoder, where a quantizer with an adjustable mask number $m$ is integrated.

\begin{figure}[tbp]
	\centering
	\includegraphics[width=0.96\linewidth]{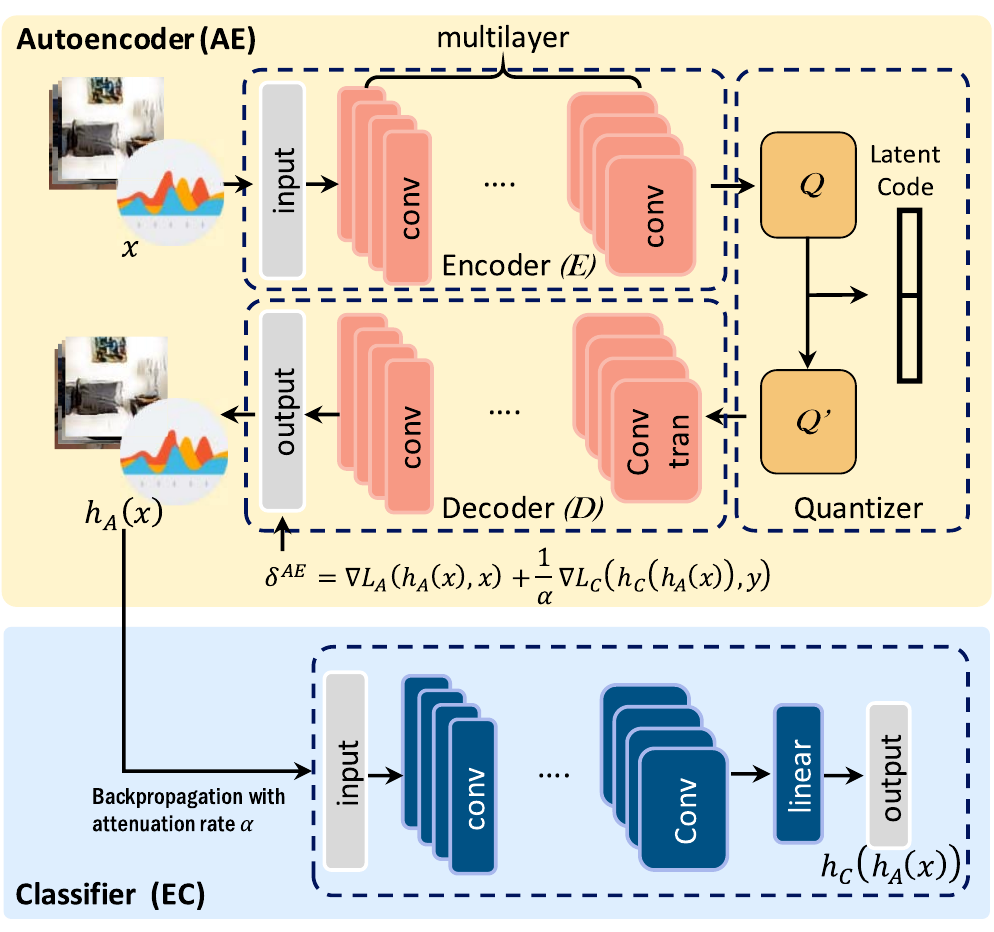}
  \caption{The architecture of CDC model. One-dimensional convolution for audios, two-dimensional for images. Top: The autoencoder with a quantization module. Except for the last convolution of $E$ and $D$, each convolution and transposed convolution is followed by a leakyReLU activation function. Mean squared error is used as a measure of distortion during training. Bottom: The elementary classifier with convolutions followed by ReLU activation functions, and is interspersed with some pooling layers.}
  \label{net_framework-pic}
\end{figure}

As shown in Figure~\ref{net_framework-pic}, our compression autoencoder contains three major components:
the encoder $E$, the decoder $D$, and the quantizer $Q$ and $Q'$.
Their definitions (as specifically designed mappings) are as follows:
\begin{gather*}
  E: \mathbb{R}^{N}\rightarrow \left( -1,\, 1\right)^{M},\quad D: \left( -1,\, 1\right)^{M}\rightarrow   \mathbb{R}^{N},\\
  Q: \left( -1,\, 1\right)^{M}\rightarrow \left [ -2^{4-m}+1, \ldots , -1, 0, 1, \ldots , 2^{4-m} \right ]^{M},\\
  Q':\left [ -2^{4-m}+1, \ldots , -1, 0, 1,\ldots , 2^{4-m} \right ]^{M}\rightarrow \left ( -1,\, 1 \right )^{M}.
\end{gather*}

In our compression autoencoder, $E$ maps $N$-dimensional input data into $M$-dimensional latent codes, where the range of each output dimension is $\left( -1,\, 1\right)$.
Correspondingly, $D$ is responsible for mapping latent codes into restored data.
In addition, $Q$ maps latent codes from $E$ into $M$-dimensional tuples, where each dimension is an integer within the range of $[-2^{4-m}+1,2^{4-m}]$ that can be efficiently stored with bits.
Here, $m$ represents the mask number of $Q$.
Correspondingly, $Q'$ restores quantized tuples as specific intervals containing pre-quantized values.

The stochastic quantization process with a uniformly distributed random noise $\varepsilon$ is conducted as follows:
\begin{gather*}
  Q\left ( x \right ) =  \left \lceil x\cdot 2^{4-m}+\varepsilon   \right \rceil_{\left ( -2^{4-m},\, 2^{4-m} \right )},\\
  Q'\left ( x\right ) = \left ( x -0.5\right )/2^{4-m}.
\end{gather*}
Here, $Q\left ( x \right )$ represents the process of quantized compression,
$Q'\left ( x\right )$ represents the process of quantized compression and reduction,
and $\left \lceil \, \right \rceil_{\left ( a,\, b \right )}$ is the ceiling rounding operation with $a$ and $b$ as lower and upper bounds, respectively. Specifically, we use $\varepsilon \in \left ( -0.5,\, 0.5 \right )$ to constrain the output of $E$, while the derivative of quantization is replaced with the derivative of the expectation. In this way, we can use the regularization provided by the stochastic quantization to cleanly backpropagate gradients through the quantizer, reduce the over-fitting caused by the fluctuation of quantized values, and achieve a better generalization capability~\cite{theis2017lossy}.

In the discussion above, the mask number $m\in\{0,1,2,3,4\}$ is treated as a \textit{compression/restoration hyperparameter}, which is used to adjust the ratio of zoom in/out before rounding for quantized compression.
Different mask numbers lead to differences in terms of bandwidth consumption and reconstruction error.
Specifically, with a lower $m$, a fewer lower numbers are truncated, and quantized data have a higher precision.
To obtain stable inputs for $D$, the addition of $\varepsilon$ allows outputs of $E$ to be better partitioned, which means that the output of $E$ is constrained between two adjacent integers during training.

It’s worth noting that on the basis of quantization, the entropy coding method, which could further improve the compression ratio, can be easily integrated into our framework.

\subsection{Discussions} \label{subsec:discussion}
\paragraph{Applying CDC in inference.} It is promising to extend the application of CDC in edge-cloud collaborative DNN inference.
Specifically, for CDC, EC on edges can be treated as early exits of inference results~\cite{teerapittayanon2017distributed,li2018edge}, which can significantly reduce the inference latency, while AC on the Cloud can provide more credible inference results with higher bandwidth consumption and inference latency.
Besides, results on edges can also assist the inference on the Cloud, where ensemble learning can be adopted to enhance inference accuracy.

\paragraph{Distributed training.} For practical applications, it is promising to combine CDC with other distributed algorithms to achieve efficient training.
For example, in the multi-edge scenario, if data on distributed edge nodes are i.i.d., collaborative training can be conducted, where data parallel methods~\cite{li2014scaling,abadi2016tensorflow,hardy2017distributed} can be adopted to improve the training speed on edges. And if data are not i.i.d., the federated learning method~\cite{kairouz2019advances} can also be adopted for diverse application.

\section{Performance Evaluation} \label{sec:eva}
In this section, we conducted extensive experiments to evaluate the performance of CDC on reducing the bandwidth consumption while preserving model accuracy for effective edge-cloud collaborative DNN training.
Our experimental methodology and results are as follows.

\subsection{System Architecture and Classification Models}
In our experiments, we constructed an edge-cloud simulator with a pair of cloud and edge servers.
Specifically, we adopted ResNeXt-29 (32x4d)~\cite{xie2017aggregated} as the advanced classifier (AC) on the Cloud for image classification, depth one-dimensional convolutional network for audio classification.
An autoencoder (AE) and an elementary classifier (EC) with the structure in Figure~\ref{net_framework-pic} were deployed on the edge server.
The model training was conducted using \texttt{PyTorch}.

\subsection{Datasets for Classification}
For a comprehensive evaluation, we adopted two different image datasets and two different audio datasets to validate the effectiveness of our approach on practical classification tasks.

\textbf{The LSUN}~\cite{yu2015lsun} contains 10 classes of images of real-life scenes. Specifically, we selected 10,000 images for each class as the training set on the edge server, and we used the test set provided by LSUN as our test set on the Cloud. 

\textbf{The Vehicles} contains images of 10 classes of vehicles selected from the imagenet dataset~\cite{deng2009imagenet}, where 12,000 images were deployed on the edge server as the training set, and 1,000 images were deployed on the Cloud as the test set. 

\textbf{The Urbansound8K}~\cite{salamon2014dataset} consists of 10 classes of audio data in .wav format. We used the preset folders 1-8 as the training set, folders 9, 10 as the test set. 

\textbf{The FSDD}~\cite{jackson2018jakobovski} is an audio dataset with 1,500 recordings in .wav format, corresponding to spoken digits ($0 \sim 9$) from 3 different speakers, where each digit is recited 50 times. We used the official training set and test set division.

\subsection{Effectiveness of Classification Guidence}
In this set of experiments, we investigated the impact of the classification guidance from the elementary classifier (EC) on the performance of both autoencoder (AE) and advanced classifier (AC).
Specifically, we used a fixed quantizer mask number $m=4$ that corresponds to 1 Bits-Per-Pixel~(BPP) image compression ratio, 13.5:1 data compression ratio for Urbansound8K, and 12:1 data compression ratio for FSDD.
We developed multiple CDC models (\emph{i.e.} AE + EC) with different attenuation rates and evaluated their capabilities on preserving both raw data features and the classification accuracy of AC.
Each CDC model was trained three times, and the one with the highest validating accuracy was selected for the result demonstration.
For each CDC model above, AC was trained three times, and, similarly, the one with the highest validating accuracy was selected for the result demonstration.

\subsubsection{Feature Preservation}
\begin{figure}[t]
  \centering
  \subfigure[$\alpha=8$.]{
    \begin{minipage}[t]{0.166\linewidth}
    \centering
    \includegraphics[width=1\linewidth]{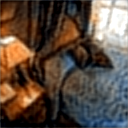}
    \includegraphics[width=1\linewidth]{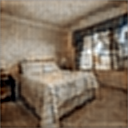}
    \end{minipage}%
  }%
  \hspace{-0.1in}
  \subfigure[$12$.]{
    \begin{minipage}[t]{0.167\linewidth}
    \centering
    \includegraphics[width=1\linewidth]{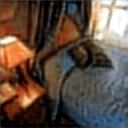}
    \includegraphics[width=1\linewidth]{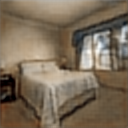}
    \end{minipage}%
  }%
  \hspace{-0.1in}
  \subfigure[$24$.]{
    \begin{minipage}[t]{0.167\linewidth}
    \centering
    \includegraphics[width=1\linewidth]{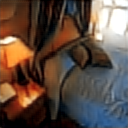}
    \includegraphics[width=1\linewidth]{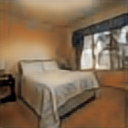}
    \end{minipage}%
  }%
  \hspace{-0.1in}
  \subfigure[$48$.]{
    \begin{minipage}[t]{0.167\linewidth}
    \centering
    \includegraphics[width=1\linewidth]{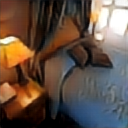}
    \includegraphics[width=1\linewidth]{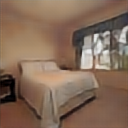}
    \end{minipage}
  }%
  \hspace{-0.12in}
  \subfigure[$256$.]{
    \begin{minipage}[t]{0.167\linewidth}
    \centering
    \includegraphics[width=1\linewidth]{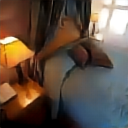}
    \includegraphics[width=1\linewidth]{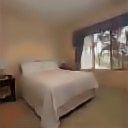}
    \end{minipage}
  }%
  \hspace{-0.12in}
  \subfigure[no EC.]{
    \begin{minipage}[t]{0.167\linewidth}
    \centering
    \includegraphics[width=1\linewidth]{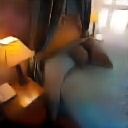}
    \includegraphics[width=1\linewidth]{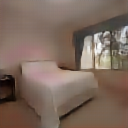}
    \end{minipage}
  }%
  \caption{Visual comparison of CDC compression results with different attenuation rates (with 1 BPP compression ratio).}
  \label{attenua-pic}
\end{figure}

We first discuss the impact of classification guidance attenuation rate on CDC's capability of preserving raw data features.
Figure~\ref{attenua-pic} illustrates several image outputs of CDC models with different attenuation rates for a visual comparison.
It is obvious that \textit{CDC focuses on raw image details more rigidly when a lower attenuation rate is adopted}, which proves the viability of integrating the training of EC as an auxiliary task for the training of AE.
One thing that should be noted is that low attenuation rates lead to severe distortions of raw data since optimization targets of EC and AE are not completely consistent, and insufficiently attenuated guidance will hinder CDC's capability on restoring data.

\subsubsection{Classification Accuracy}
\begin{figure*}[t]
  \centering
  \subfigure[On LSUN.]{
    \begin{minipage}[t]{0.25\linewidth}
    \centering
    \includegraphics[width=1\linewidth]{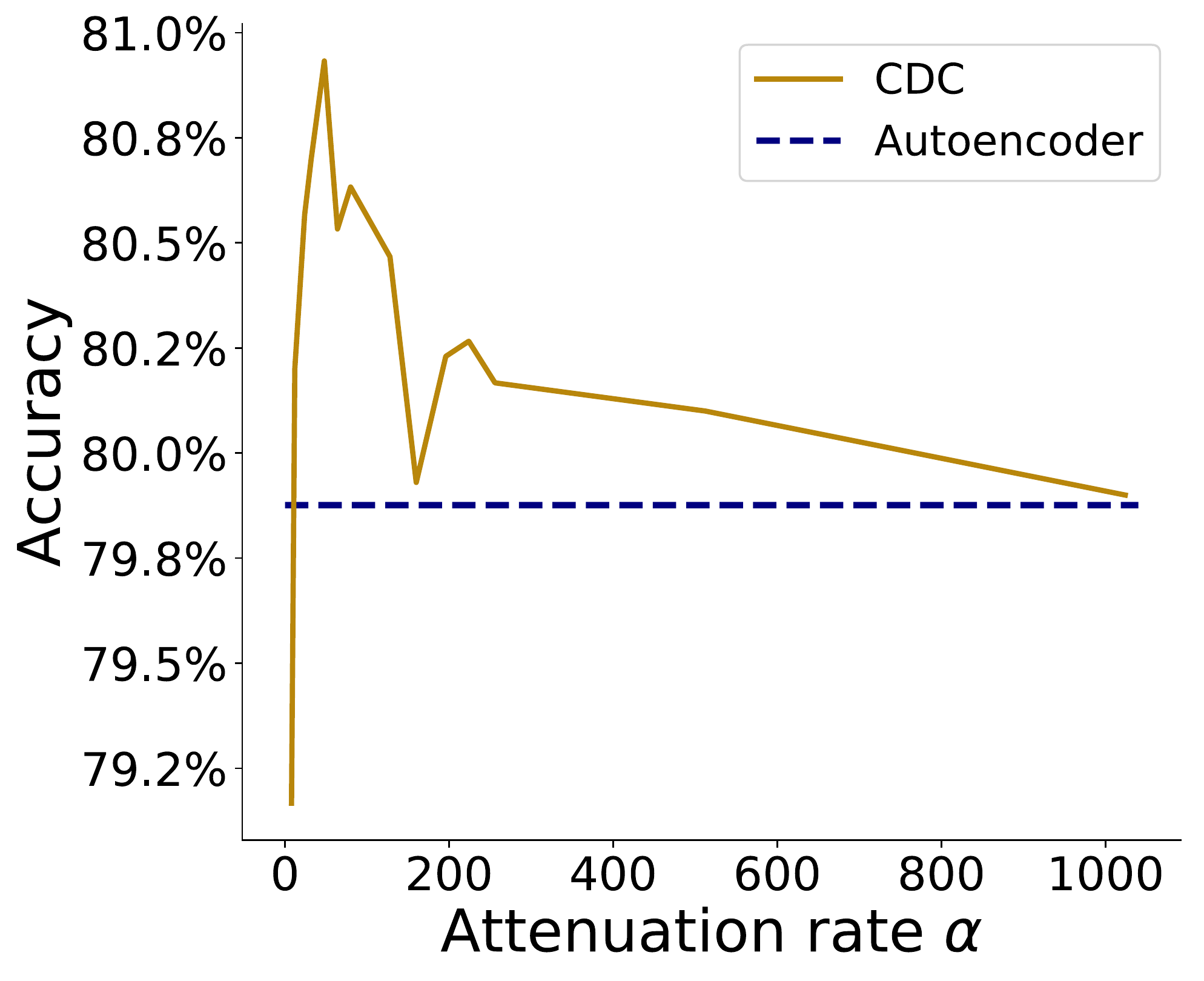}
    \end{minipage}%
  }%
  \hspace{-0.12in}
  \subfigure[On Vehicles.]{
    \begin{minipage}[t]{0.25\linewidth}
    \centering
    \includegraphics[width=1\linewidth]{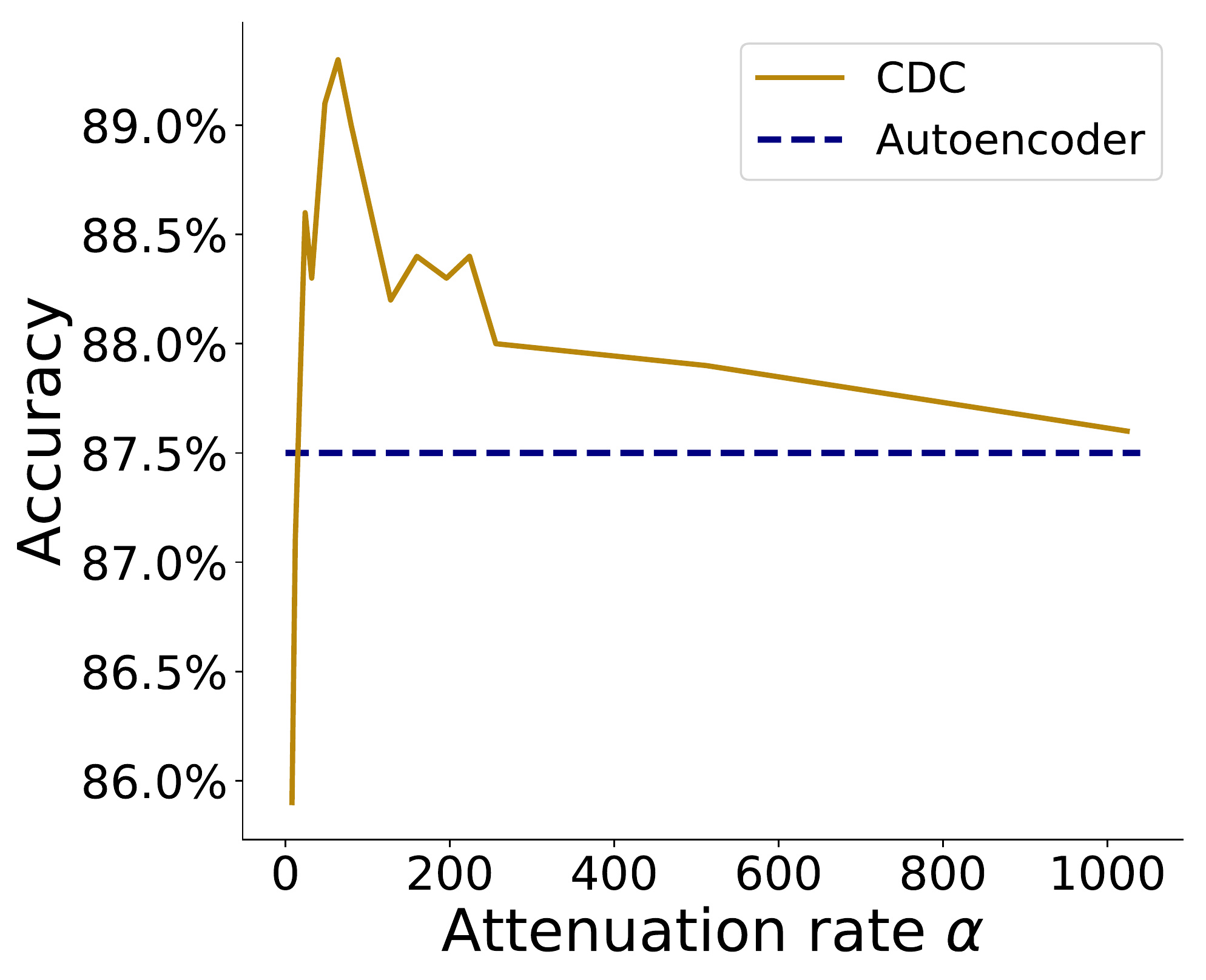}
    \end{minipage}%
  }%
  \hspace{-0.12in}
  \subfigure[On Urbansound8K.]{
    \begin{minipage}[t]{0.25\linewidth}
    \centering
    \includegraphics[width=1\linewidth]{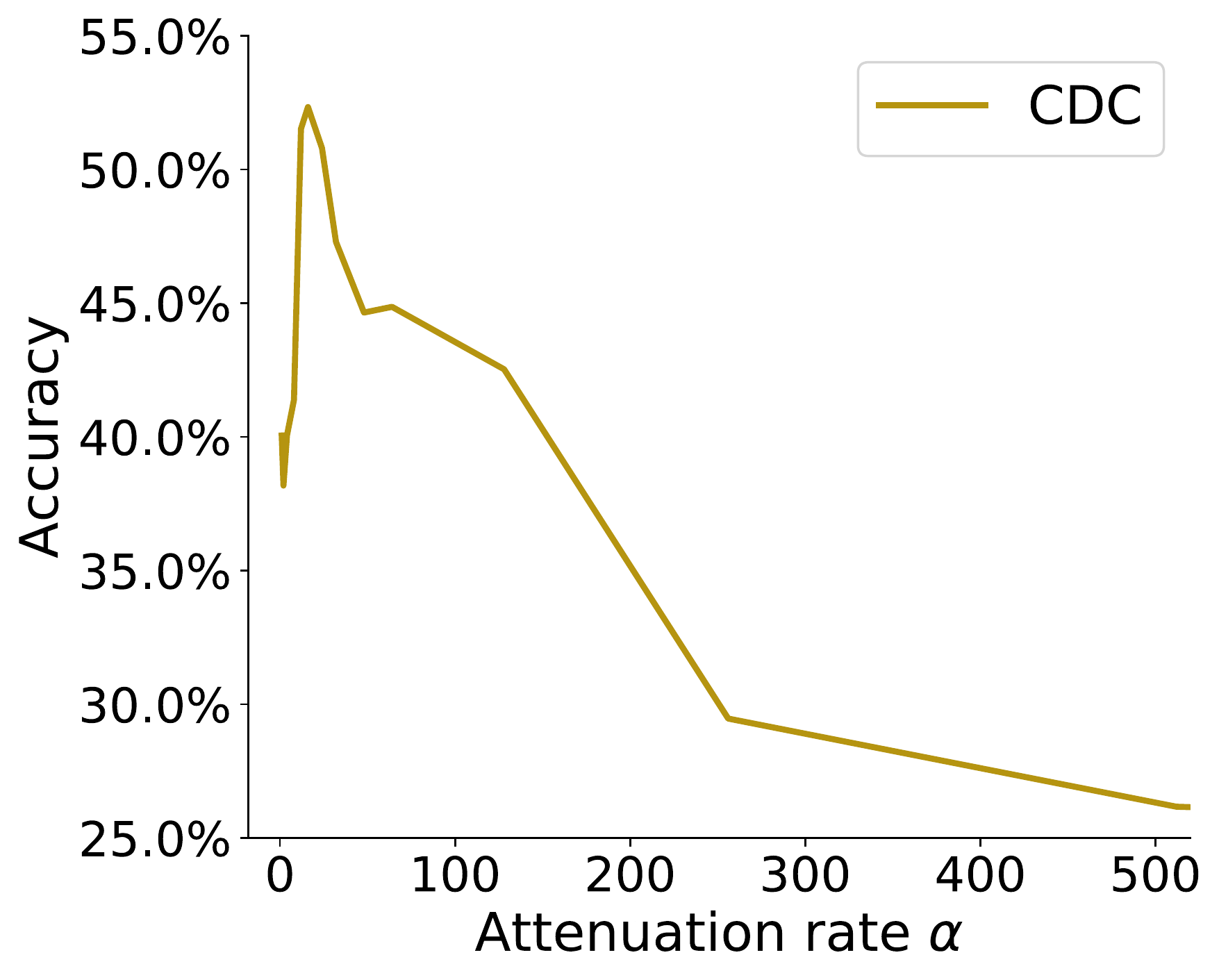}
    \end{minipage}
  }%
  \hspace{-0.12in}
  \subfigure[On FSDD.]{
    \begin{minipage}[t]{0.25\linewidth}
    \centering
    \includegraphics[width=1\linewidth]{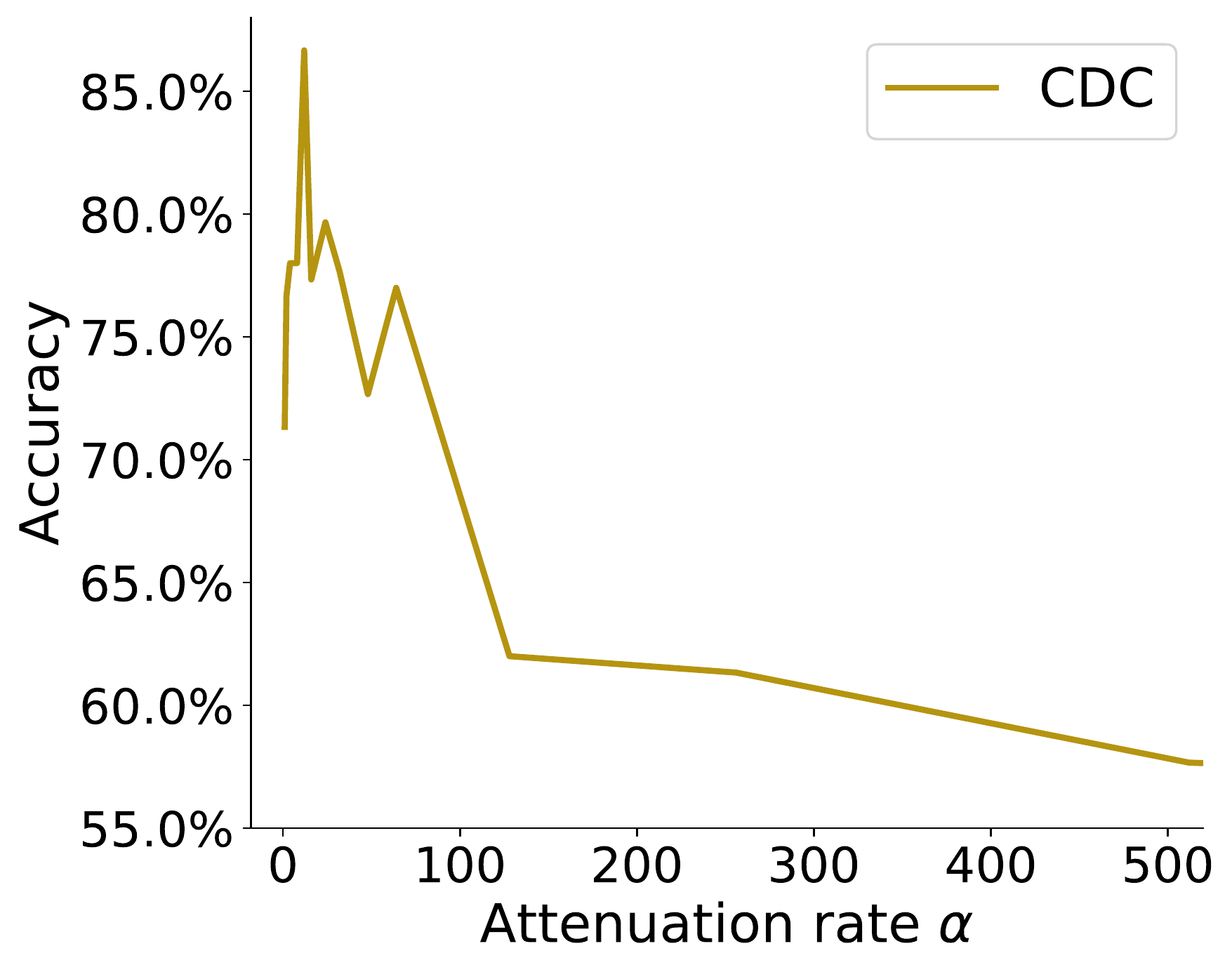}
    \end{minipage}
  }%
  \caption{Impact of classification guidance attenuation rate on AC accuracy.}
  \label{attenua_result-pic}
\end{figure*}

We now discuss the impact of classification guidance attenuation rate on the classification accuracy of AC.
The AC accuracy with CDC models using different attenuation rates is demonstrated in Figure~\ref{attenua_result-pic}, where the AC accuracy with traditional Autoencoder is also depicted for comparison.
Obviously, on all adopted datasets, the CDC model with a proper attenuation rate (\emph{i.e.} from our experiments, around 48 for images, 16 for audios) manages to achieve a significantly higher AC accuracy compared with the scenario with traditional Autoencoder. 
In particular, AC trained on the audios compressed and restored by the traditional autoencoder do not converge, while the data compressed and restored by CDC can effectively support the classification on the Cloud.
Such a result indicates that the integration of classification guidance with a proper attenuation rate can obviously enhance the capability of an autoencoder on selectively preserving raw data features that are critical to the training of high-accuracy advanced classifiers on the Cloud.

For a further validation, we compared the performance of the optimal CDC model on preserving the AC accuracy with that of traditional Autoencoder.
Specifically, we separately trained AC with our preprocessed four datasets without any compression as baselines.
Then, with a fixed compression ratio (\emph{i.e.} $m=4$), AC was trained based on the restored datasets from traditional autoencoder and CDC, respectively.
As shown in Table~\ref{CDC-table1}, CDC outperforms its comparative under the same compression ratio requirement.

\begin{table}[tbp]
  \centering
  \begin{tabular}{lcc}
    \toprule
    Methods & Accuracy (L / V)(\%) & Loss (L / V)(\%)\\
    \cmidrule(lr){2-3}
     & Accuracy (U / F)(\%) & Loss (U / F)(\%)\\
    \midrule
    Baseline & 81.99 / 90.50 & 0 / 0 \\
    \cmidrule(lr){2-3}
     & 62.37 / 88.33 & 0 / 0 \\
    \cmidrule(lr){1-3}
    AE & 79.87 / 87.50 & 2.12 / 3.00 \\
    \cmidrule(lr){2-3}
     & 12.44 / 10.67 & 49.93 / 77.66 \\
     \cmidrule(lr){1-3}
    CDC & 80.93 / 89.10 & 1.06 / 1.40 \\
    \cmidrule(lr){2-3}
    & 52.33 / 86.67 & 10.04 / 1.66 \\
    \bottomrule
  \end{tabular}
  \caption{AC accuracy with traditional AE and CDC (with attenuation rate $\alpha=48$ for image datasets, $\alpha=16$ for audio). L, V, U and F respectively stand for LSUN, Vehicles, Urbansound8K and FSDD.}
  \label{CDC-table1}
\end{table}

\subsection{Tradeoff between Compression Ratio and Classification Accuracy}
In this set of experiments, we investigated CDC's capability on achieving the tradeoff between raw data compression (to reduce bandwidth consumption) and the classification accuracy during the edge-cloud collaborative training process.
Specifically, with a fixed classification guidance attenuation rate (\emph{i.e.} $\alpha = 48$ in image datasets, $\alpha = 16$ in audio datasets), we developed multiple CDC models with different compression ratios by adjusting the mask number $m$ of AE.

Table~\ref{comp_rate-table} illustrates the AC accuracy of CDC models with different compression ratios as well as the corresponding accuracy loss trained with the preprocessed LSUN dataset (\emph{i.e.} 81.99\%)\footnote{We explicitly present the result of LSUN with images of real-life scenes, where the performance of CDC is relatively representative on such large-scale dataset, for a more convincing discussion.}.
As we can see, under the experimental setting, compared with training AC with the original dataset (preprocessed LSUN), our approach manages to reduce the bandwidth consumption by 93.7\% with an accuracy loss of only 1.06\%.
According to Table~\ref{comp_rate-table}, in general, the accuracy loss of CDC increases with the intensity of raw data compression.
It should be noted that with a 4~BPP compression ratio, CDC has a higher classification accuracy compared with the baseline AC.
We believe that AE with classification guidance from EC on the edge manages to assist the training of AC on the Cloud by successfully preserving classification related features. 
Considering that the accuracy loss increases in the 5~BPP scenario, we would believe that too high the image restoration capability may weaken the emphasis of classification related information brought by the guidance.

\begin{table}[tbp]
  \centering
  \begin{tabular}{lccc}
    \toprule
    BPP & Bandwidth reduction & Accuracy  & Accuracy loss\\
    \midrule
      1 & 93.7\% & 80.93\% & 1.06\%       \\
      2 & 87.4\% & 81.14\% & 0.85\%       \\
      3 & 81.1\% & 81.67\% & 0.32\%       \\
      4 & 74.8\% & 82.05\% & -0.06\%      \\
      5 & 68.5\% & 81.92\% & 0.07\%       \\
      \bottomrule
  \end{tabular}
  \caption{Relation between AE compression ratio and AC accuracy.}
  \label{comp_rate-table}
\end{table}

\begin{figure}[t]
  \centering
  \begin{tabular}{c}
    \begin{minipage}[t]{0.80\linewidth}
      \includegraphics[width=1\linewidth]{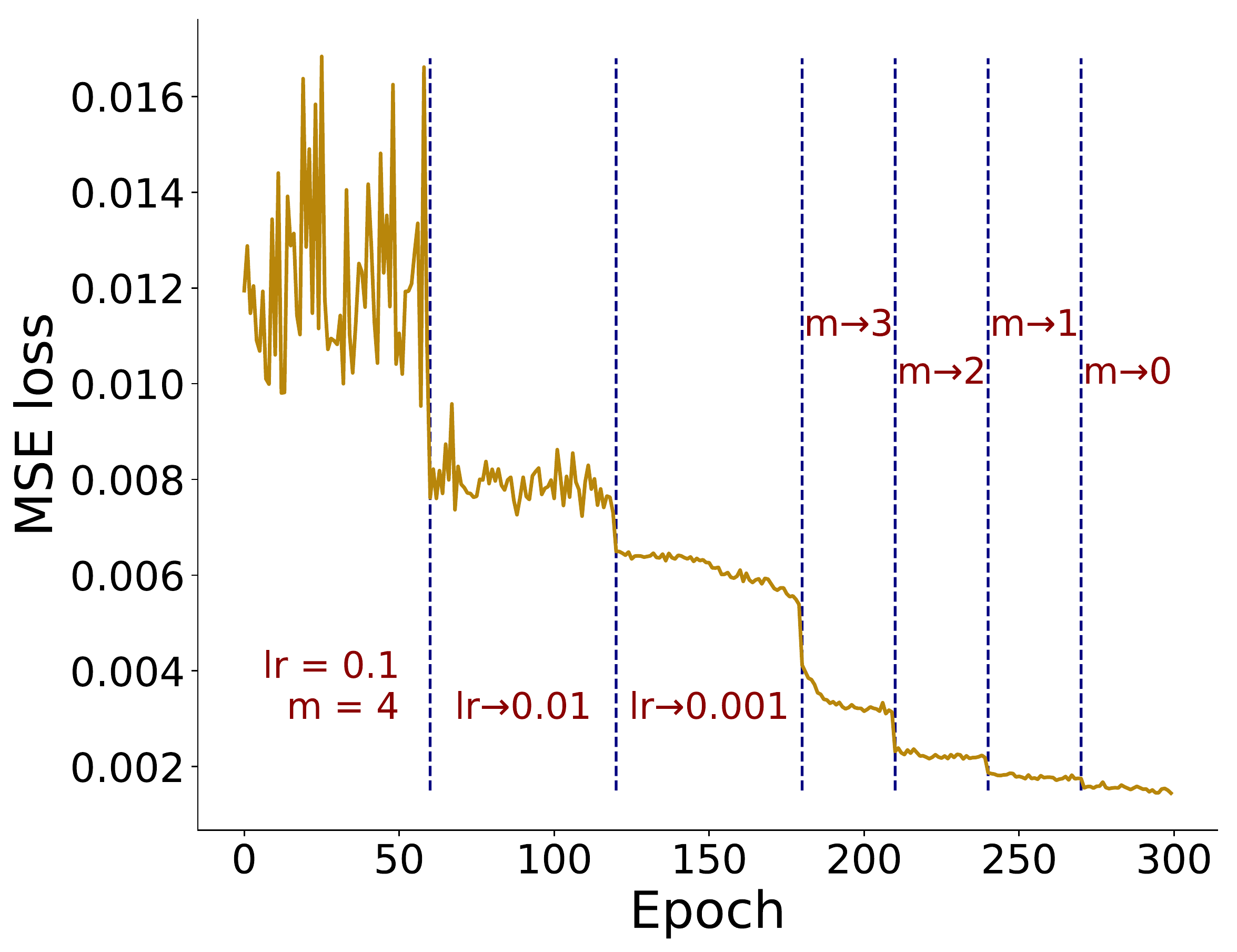}
    \end{minipage}%
  \end{tabular}

  \caption{Impact of the compression ratio switching on CDC.}
  \label{change_mask-pic}
\end{figure}

To further evaluate CDC's capability on dealing with compression ratio switches (according to the dynamic network bandwidth status in Figure~\ref{framework-pic}), we changed $m$ in different manners and fine-tuned our model with a low learning rate.
The variation of MSE under different learning rates and masks are illustrated in Figure~\ref{change_mask-pic}.
As we can see, our model manages to rapidly converge after compression ratio adjustments.
It should be noted that when the compression ratio decreases, the training curve does not oscillate and the convergence is quite smooth.

\section{Conclusion}
In this paper, we present Classification Driven Compression (CDC) for effective edge-cloud collaborative classifier training, where the tradeoff between network bandwidth consumption and classification accuracy can be achieved.
Specifically, we develop a classification-guided autoencoder to compress training data while preserving critical features for the training of high-accuracy classifier on the Cloud, and a quantization based compression method for the bandwidth-accuracy tradeoff.
We conducted extensive experiments to evaluate the performance of CDC.
For the edge-cloud collaborative training of a DNN classifier (\emph{i.e.} ResNext-29), compared with that based on raw data, CDC manages to consume 14.9$\times$ less bandwidth by introducing an accuracy loss no more than 1.06\%;
compared with DNN training with data compressed by AE without guidance, CDC introduces at least 100\% lower accuracy loss.

\bibliographystyle{named}
\bibliography{Reference}

\begin{thebibliography}{}

\bibitem[\protect\citeauthoryear{Abadi \bgroup \em et al.\egroup
  }{2016}]{abadi2016tensorflow}
Mart{\'\i}n Abadi, Ashish Agarwal, Paul Barham, Eugene Brevdo, et~al.
\newblock Tensorflow: Large-scale machine learning on heterogeneous distributed
  systems.
\newblock {\em arXiv preprint arXiv:1603.04467}, 2016.

\bibitem[\protect\citeauthoryear{Chen and Ran}{2019}]{8763885}
Jiasi Chen and Xukan Ran.
\newblock Deep learning with edge computing: A review.
\newblock {\em Proc. IEEE}, 107(8):1655--1674, 2019.

\bibitem[\protect\citeauthoryear{Cipolla \bgroup \em et al.\egroup
  }{2018}]{cipolla2018multi}
Roberto Cipolla, Yarin Gal, and Alex Kendall.
\newblock Multi-task learning using uncertainty to weigh losses for scene
  geometry and semantics.
\newblock In {\em Proc. of IEEE CVPR}, pages 7482--7491, 2018.

\bibitem[\protect\citeauthoryear{Cisco}{2016}]{networking2016cisco}
Cisco.
\newblock Cisco global cloud index: Forecast and methodology, 2016--2021.
\newblock {\em Cisco White paper}, 2016.

\bibitem[\protect\citeauthoryear{Deng \bgroup \em et al.\egroup
  }{2009}]{deng2009imagenet}
Jia Deng, Wei Dong, Richard Socher, Li-Jia Li, Kai Li, and Li~Fei-Fei.
\newblock Imagenet: A large-scale hierarchical image database.
\newblock In {\em Proc. IEEE CVPR}, pages 248--255, 2009.

\bibitem[\protect\citeauthoryear{Google}{2018}]{googleEI}
Google.
\newblock https://cloud.google.com/blog/
  products/gcp/bringing-intelligence-edge-cloud-iot, 2018.

\bibitem[\protect\citeauthoryear{Hardy \bgroup \em et al.\egroup
  }{2017}]{hardy2017distributed}
Corentin Hardy, Erwan Le~Merrer, and Bruno Sericola.
\newblock Distributed deep learning on edge-devices: feasibility via adaptive
  compression.
\newblock In {\em Proc. IEEE NCA}, pages 1--8. IEEE, 2017.

\bibitem[\protect\citeauthoryear{Jackson \bgroup \em et al.\egroup
  }{2018}]{jackson2018jakobovski}
Zohar Jackson, Csar Souza, Jason Flaks, and H~Nicolas.
\newblock Jakobovski/free-spoken-digit-dataset v1. 0.7, 2018.

\bibitem[\protect\citeauthoryear{Kairouz \bgroup \em et al.\egroup
  }{2019}]{kairouz2019advances}
Peter Kairouz, H~Brendan McMahan, Brendan Avent, Aur{\'e}lien Bellet, Mehdi
  Bennis, Arjun~Nitin Bhagoji, Keith Bonawitz, Zachary Charles, Graham Cormode,
  Rachel Cummings, et~al.
\newblock Advances and open problems in federated learning.
\newblock {\em arXiv preprint arXiv:1912.04977}, 2019.

\bibitem[\protect\citeauthoryear{Le \bgroup \em et al.\egroup
  }{2018}]{le2018supervised}
Lei Le, Andrew Patterson, and Martha White.
\newblock Supervised autoencoders: Improving generalization performance with
  unsupervised regularizers.
\newblock In {\em Proc. NeurIPS}, pages 107--117, 2018.

\bibitem[\protect\citeauthoryear{Li \bgroup \em et al.\egroup
  }{2014}]{li2014scaling}
Mu~Li, David~G Andersen, Jun~Woo Park, Alexander~J Smola, et~al.
\newblock Scaling distributed machine learning with the parameter server.
\newblock In {\em Proc. OSDI}, pages 583--598, 2014.

\bibitem[\protect\citeauthoryear{Li \bgroup \em et al.\egroup
  }{2018a}]{li2018edge}
En~Li, Zhi Zhou, and Xu~Chen.
\newblock Edge intelligence: On-demand deep learning model co-inference with
  device-edge synergy.
\newblock In {\em Proc. Workshop on Mobile Edge Communications}, pages 31--36.
  ACM, 2018.

\bibitem[\protect\citeauthoryear{Li \bgroup \em et al.\egroup
  }{2018b}]{li2018learning}
He~Li, Kaoru Ota, and Mianxiong Dong.
\newblock Learning iot in edge: deep learning for the internet of things with
  edge computing.
\newblock {\em IEEE Network}, 32(1):96--101, 2018.

\bibitem[\protect\citeauthoryear{Li \bgroup \em et al.\egroup
  }{2018c}]{li2018learning_}
Mu~Li, Wangmeng Zuo, Shuhang Gu, Debin Zhao, and David Zhang.
\newblock Learning convolutional networks for content-weighted image
  compression.
\newblock In {\em Proc. of IEEE CVPR}, pages 3214--3223, 2018.

\bibitem[\protect\citeauthoryear{Liu \bgroup \em et al.\egroup
  }{2016}]{liu2016algorithm}
Tongliang Liu, Dacheng Tao, Mingli Song, and Stephen~J Maybank.
\newblock Algorithm-dependent generalization bounds for multi-task learning.
\newblock {\em IEEE PAMI}, 39(2):227--241, 2016.

\bibitem[\protect\citeauthoryear{Mell \bgroup \em et al.\egroup
  }{2011}]{mell2011nist}
Peter Mell, Tim Grance, et~al.
\newblock The nist definition of cloud computing.
\newblock 2011.

\bibitem[\protect\citeauthoryear{Microsoft}{2019}]{msEI}
Microsoft.
\newblock https://azure.microsoft.com/ en-us/overview/future-of-cloud, 2019.

\bibitem[\protect\citeauthoryear{Salamon \bgroup \em et al.\egroup
  }{2014}]{salamon2014dataset}
Justin Salamon, Christopher Jacoby, and Juan~Pablo Bello.
\newblock A dataset and taxonomy for urban sound research.
\newblock In {\em Proc. ACM ICME}, pages 1041--1044, 2014.

\bibitem[\protect\citeauthoryear{Satyanarayan}{2017}]{satyanarayanan2017emergence}
Mahadev Satyanarayan.
\newblock The emergence of edge computing.
\newblock {\em Computer}, 50(1):30--39, 2017.

\bibitem[\protect\citeauthoryear{Shaham and
  Michaeli}{2018}]{Shaham2018Deformation}
Tamar~Rott Shaham and Tomer Michaeli.
\newblock Deformation aware image compression.
\newblock In {\em Proc. of IEEE CVPR}, 2018.

\bibitem[\protect\citeauthoryear{Shi and Dustdar}{2016}]{shi2016promise}
Weisong Shi and Schahram Dustdar.
\newblock The promise of edge computing.
\newblock {\em Computer}, 49(5):78--81, 2016.

\bibitem[\protect\citeauthoryear{Smith \bgroup \em et al.\egroup
  }{2017}]{smith2017federated}
Virginia Smith, Chao-Kai Chiang, Maziar Sanjabi, and Ameet~S Talwalkar.
\newblock Federated multi-task learning.
\newblock In {\em Proc. of NeurIPS}, pages 4424--4434, 2017.

\bibitem[\protect\citeauthoryear{Tang \bgroup \em et al.\egroup
  }{2018}]{tang2018communication}
Hanlin Tang, Shaoduo Gan, Ce~Zhang, Tong Zhang, and Ji~Liu.
\newblock Communication compression for decentralized training.
\newblock In {\em Proc. NeurIPS}, pages 7652--7662, 2018.

\bibitem[\protect\citeauthoryear{Teerapittayanon \bgroup \em et al.\egroup
  }{2017}]{teerapittayanon2017distributed}
Surat Teerapittayanon, Bradley McDanel, and HT~Kung.
\newblock Distributed deep neural networks over the cloud, the edge and end
  devices.
\newblock In {\em Proc. IEEE ICDCS}, pages 328--339, 2017.

\bibitem[\protect\citeauthoryear{Theis \bgroup \em et al.\egroup
  }{2017}]{theis2017lossy}
Lucas Theis, Wenzhe Shi, Andrew Cunningham, and Ferenc Husz{\'a}r.
\newblock Lossy image compression with compressive autoencoders.
\newblock {\em ICLR}, 2017.

\bibitem[\protect\citeauthoryear{Xie \bgroup \em et al.\egroup
  }{2017}]{xie2017aggregated}
Saining Xie, Ross Girshick, Piotr Doll{\'a}r, Zhuowen Tu, and Kaiming He.
\newblock Aggregated residual transformations for deep neural networks.
\newblock In {\em Proc. of IEEE CVPR}, pages 1492--1500, 2017.

\bibitem[\protect\citeauthoryear{Yu \bgroup \em et al.\egroup
  }{2015}]{yu2015lsun}
Fisher Yu, Ari Seff, Yinda Zhang, Shuran Song, et~al.
\newblock Lsun: Construction of a large-scale image dataset using deep learning
  with humans in the loop.
\newblock {\em arXiv preprint arXiv:1506.03365}, 2015.

\bibitem[\protect\citeauthoryear{Zhou \bgroup \em et al.\egroup
  }{2018}]{zhou2018variational}
Lei Zhou, Chunlei Cai, Yue Gao, Sanbao Su, and Junmin Wu.
\newblock Variational autoencoder for low bit-rate image compression.
\newblock In {\em Proc. IEEE CVPR}, pages 2617--2620, 2018.

\bibitem[\protect\citeauthoryear{Zhou \bgroup \em et al.\egroup
  }{2019}]{8736011}
Zhi Zhou, Xu~Chen, En~Li, Liekang Zeng, Ke~Luo, and Junshan Zhang.
\newblock Edge intelligence: Paving the last mile of artificial intelligence
  with edge computing.
\newblock {\em Proc. IEEE}, 107(8):1738--1762, 2019.

\end{thebibliography}

\end{document}